\documentclass{article}
\usepackage{ijcai22}

\usepackage{times}
\usepackage{soul}
\usepackage{url}
\usepackage[hidelinks]{hyperref}
\usepackage[utf8]{inputenc}
\usepackage[small]{caption}
\usepackage{graphicx}
\usepackage{amsmath}
\usepackage{amsthm}
\usepackage{booktabs}
\usepackage{algorithm}
\usepackage{algorithmic}
\usepackage{multirow}
\usepackage{stfloats}
\usepackage[justification=centering]{caption}

\title{Learn to Cluster Faces with Better Subgraphs}

\author{
Yuan Cao$^1$\and
Di Jiang$^1$\and
Guanqun Hou$^2$\and
Fan Deng$^2$\and
Xinjia Chen$^2$\and
Qiang Yang$^1$
\affiliations
$^1$College of Electrical Engineering, Zhejiang University, Hangzhou, China\\
$^2$Hikvision Research Institute, Hangzhou, China
\emails
\{cy1998, jiang\_di, qyang\}@zju.edu.cn,
\{houguanqun, dengfan, chenxinjia\}@hikvision.com,
}
\begin{document}

\maketitle

\begin{abstract}
    Face clustering can provide pseudo-labels to the massive unlabeled face data and improve the performance of different face recognition models. The existing clustering methods generally aggregate the features within subgraphs that are often implemented based on a uniform threshold or a learned cutoff position. This may reduce the recall of subgraphs and hence degrade the clustering performance. This work proposed an efficient neighborhood-aware subgraph adjustment method that can significantly reduce the noise and improve the recall of the subgraphs, and hence can drive the distant nodes to converge towards the same centers. More specifically, the proposed method consists of two components, i.e. face embeddings enhancement using the embeddings from neighbors, and enclosed subgraph construction of node pairs for structural information extraction. The embeddings are combined to predict the linkage probabilities for all node pairs to replace the cosine similarities to produce new subgraphs that can be further used for aggregation of GCNs or other clustering methods. The proposed method is validated through extensive experiments against a range of clustering solutions using three benchmark datasets and numerical results confirm that it outperforms the SOTA solutions in terms of generalization capability.
\end{abstract}

\section{Introduction}

With the advances in deep learning technologies, face recognition and related fields have made great achievements. However, the size of high-quality face datasets has placed restrictions on the performance of face recognition models. Large-scale datasets often rely on expensive and unacceptable annotations and face clustering provides a possibility for high-quality automatic annotation. 

Owing to graph convolutional networks (GCNs), much research effort has been made in recent years. Many solutions ~\cite{L-GCN,STAR-FC,Pairwise,NDDe} have been developed based on GCNs to update features or predict linkages and made significant improvements in clustering performance. However, some outstanding challenges remain. Unlike traditional graph-related datasets, face datasets do not have an explicit graph structure. A common practice for constructing subgraphs is to calculate top-k neighbors based on cosine similarity for each node and control the noise ratio with a uniform threshold. Due to the limitation of the quality of the original features, a fixed threshold may limit the quality of the subgraphs to some extent. To overcome the shortcomings of the missing graph structures, Ada-NETs ~\cite{Ada-NETS} proposed a metric to measure the quality of subgraphs and adjusted them with learned cutoff positions. A similar module appeared in FaceMap ~\cite{FaceMap}. Although the available solutions can improve the precision of subgraphs (i.e., the proportion of the same class nodes in subgraphs), the recall (i.e., the proportion of all same class nodes) is still considered a bottleneck for further performance improvement.

\begin{figure}[tbp]
    \centering
	\includegraphics[width=0.7\linewidth]{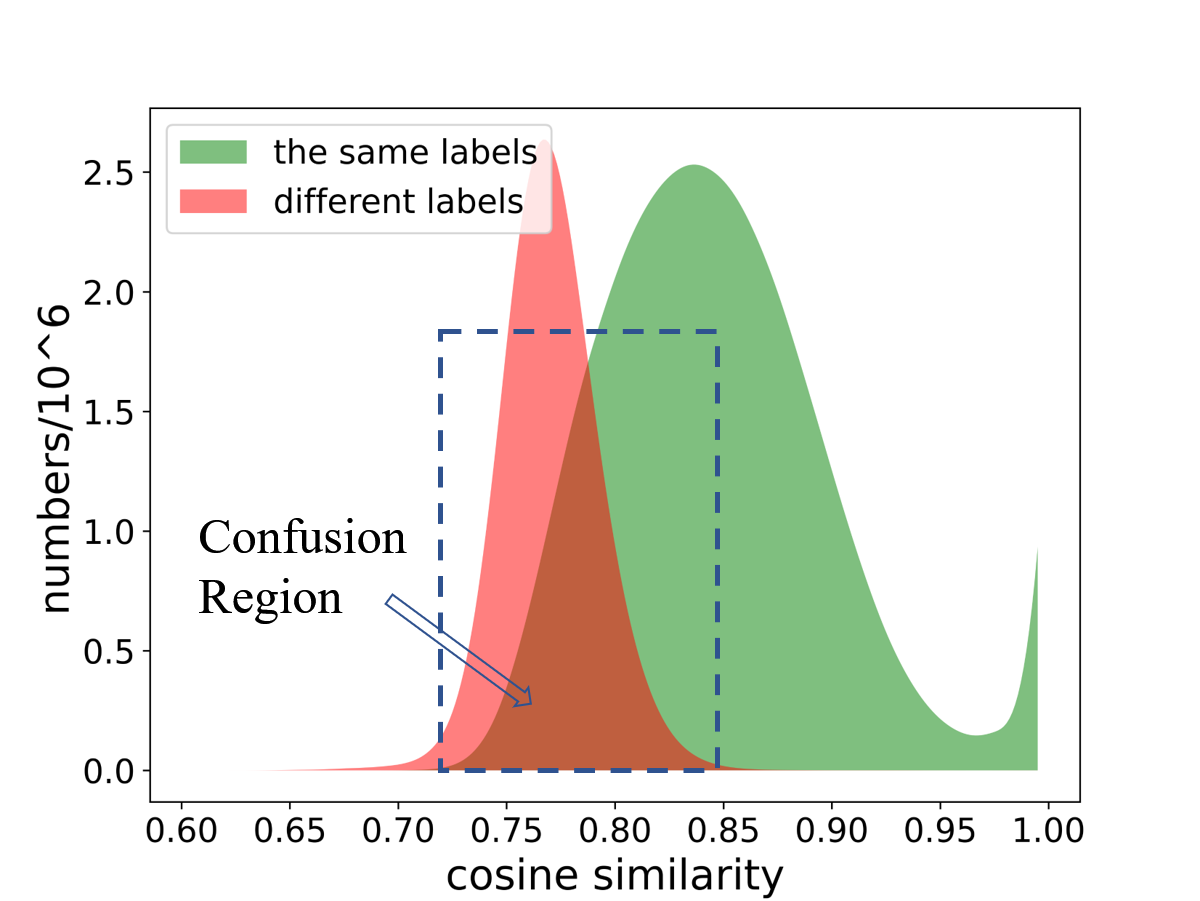}
	\caption{The kNN neighbor similarity distribution (k = 80). The confusion region is inside the blue dashed line, which contains many node pairs that cannot distinguished the connections only by cosine similarities of these features.}
\end {figure}

The precision and recall of the subgraphs have different effects during the clustering stage. Taking GCN aggregation as an example, when the precision is higher and the recall is lower, the dispersed similar nodes tend to form multiple aggregation centers, which leads to the appearance of splitting of singletons or small clusters. Conversely, dissimilar nodes tend to aggregate towards the same centers, thus making the final features lose discriminatory. Therefore, to further improve the quality of the subgraphs, both precision and recall need to be fully considered. \textbf{However, this is considered a non-trivial issue due to the insufficient representation ability of original features.} There exists a confusion region where there are a large number of nodes with different labels sharing high cosine similarities and the same class nodes with low similarities, and hence they cannot be correctly distinguished by features alone, as shown in Figure 1.

To overcome the above challenges, relying on features only is not sufficient and it needs to combine with neighborhood structure information. Neighborhood structures contain lots of information. For instance, a pair of nodes with more common neighbors is more inclined to share a similar label. In some previous studies in the field of link prediction and community detection, structures have been demonstrated that can help to improve the performance of pairwise classification. Based on these above discoveries, this work introduces a new Neighborhood-Aware Subgraph Adjustment method (NASA), which encodes enclosed subgraphs of node pairs and combines two kinds of embeddings to achieve higher performance in the subgraph adjustment stage. The effectiveness of the adjusted subgraphs with the NASA module can significantly improve the performance of GCNs and excellent scalability is demonstrated via two different clustering methods. The main contributions of this work are summarized as follows:

\begin{itemize}
    \item[$\bullet$] The effect of the precision and recall of subgraphs is extensively analyzed, especially the effect of recall on singletons.
    \item[$\bullet$] The proposed NASA module improves the precision (reducing the noisy edges in the subgraphs more effectively) and recall (increasing the number of correct edges that would have been excluded in previous methods) of the subgraphs, thus greatly improving the clustering performance of GCNs or other clustering methods.
    \item[$\bullet$] State-of-the-art performance is achieved across three mainstream face clustering benchmarks, which demonstrates a high level of generalization of the proposed methods. The enhancement for different clustering methods also demonstrates its excellent scalability.
\end{itemize}

\section{Related Work}

\paragraph{Face Clustering.} Face clustering task focuses on assigning labels to unlabeled face images to save manual labeling costs. The early face clustering algorithms usually use unsupervised clustering methods, such K-Means ~\cite{K-Means}, DBSCAN ~\cite{DBSCAN} or hierarchical clustering ~\cite{HAC}. Since these methods have strong assumptions about data distribution, they often do not work well on complex face datasets. Despite the advancements in similarity measurement and clustering techniques made in previous studies \cite{ARO,PAHC}, the performance and computational cost still fall short of the demand. With the rise and rapid development of GCN, its applications in face clustering have become a research spot. L-GCN ~\cite{L-GCN} performed link prediction with decentralized features, and DA-Net ~\cite{DANET} regarded clustering as a node classification task to update features with a BiLSTM. Two-stage networks GCN-DS ~\cite{GCN-DS} and GCN-VE ~\cite{GCN-VE} cascaded with coarse clustering and fine-tuning. STAR-FC ~\cite{STAR-FC} proposed a structure-preserved sampling strategy based on the L-GCN paradigm. Additionally, efforts ~\cite{Clusformer,FaceT} are underway to integrate Transformers into the feature aggregation stage to enhance the quality of aggregated features. Pair-Cls ~\cite{Pairwise} developed a simple yet efficient density-based link prediction approach. NDDe \& TPDi ~\cite{NDDe} were subsequently proposed to improve the performance of pairwise classification in small clusters and sparse regions, respectively. However, little work has focused on the construction of subgraphs for face clustering. Ada-NETs ~\cite{Ada-NETS} and FaceMap ~\cite{FaceMap} have proposed AND module and OD module to alleviate low-quality subgraph problems. GCN-F\&A ~\cite{GCNFA} implicitly improved the quality of subgraphs with GCN-FT and perform feature aggregation and pairwise classification sequentially. Since these works pay more attention to the precision of the subgraphs, the issue of splitting the dispersed nodes in the feature space happened. To solve this issue, additional structure embeddings should be introduced to assist the pairwise classification task, which significantly increases both the precision and the recall.

\paragraph{Link Prediction.} Conventional link prediction approaches are mainly based on the common neighbor algorithm and its extension of heuristic methods (e.g., Adamic Adar ~\cite{AA}, Jaccard ~\cite{Jaccard} or Katz ~\cite{Katz}). However, these methods have strong prior knowledge which leads to poor generalization. The authors of WLNM ~\cite{WLNM} and SEAL ~\cite{SEAL} demonstrated that incorporating enclosed subgraphs related to links into the learning stage implicitly learns the structural information around the links, which helps to improve performance and generalization. Inspired by these findings, a concise approach for incorporating structural embeddings into a pairwise classification is proposed in this work.

\section{Motivations}

In general, two elements are used in the graph structure learning solutions to determine the quality of subgraphs: necessity and adequacy. Necessity refers to the fact that subgraphs generally do not contain any task-irrelevant information, while adequacy requires that the subgraphs contain enough information for downstream tasks. Here, they can be represented by precision and recall. They play different roles in GCN aggregation. The former encourages the nearby nodes to aggregate towards an implicit center without noise disturbance, while the latter ensures that identical nodes can be contacted as widely as possible and converge to the same center in the aggregation stage. Figure 2 shows the effect of both on the clustering performance of MS1M-Part1. When the precision decreases, the complex noise disturbs the original feature distribution and causes serious performance degradation. Although the impact of recall is smaller than precision (consistent with ~\cite{Ada-NETS}), it is still not negligible. Nodes at the edges of clusters are more affected. As the recall decreases, the neighbors in subgraphs of edge nodes further reduce, making it impossible to move toward the aggregation centers and become singletons.

\begin{figure}[tbp]
    \centering
	\begin{minipage}{0.49\linewidth}
	\centering
	\includegraphics[width=4.4cm,height=3.3cm]{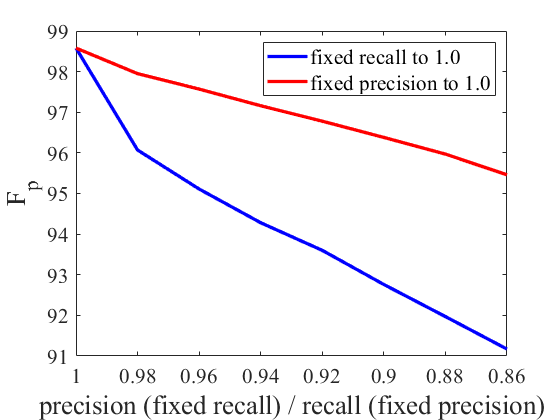}
	\caption*{(a)}
	\label{(a)}
	\end{minipage}
        \begin{minipage}{0.49\linewidth}
	\centering
	\includegraphics[width=4.4cm,height=3.3cm]{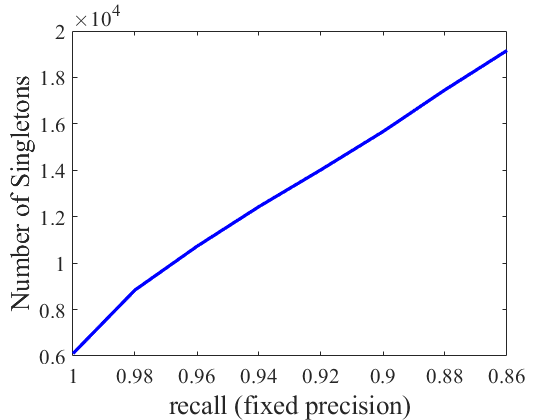}
	\caption*{(b)}
	\label{(b)}
	\end{minipage}
	\caption{Effect of the precision and recall of subgraphs on the performance of MS1M-Part1. (a) Effect of precision when fixing recall to 1.0 (blue) and effect of recall when fixing precision to 1.0 (red); (b) The trend of the number of singletons as recall decreases.}
\end {figure}

Subgraph improvement schemes have appeared in Ada-NETs and FaceMap. Despite the differences between schemes, they both raise the subgraph’s quality by obtaining neighborhood cutoff positions $k^{off}$ for all nodes. The Differences in subgraph adjustment methods are shown in Figure 3. It is demonstrated that these approaches focus more on precision and ignore the importance of recall. To improve the recall of subgraphs, it’s necessary to take all neighbors into account. An interesting finding was reported in \cite{Pairwise} that combination features can significantly improve the recall of pairwise classification in subgraphs. However, this approach may lead to a precision reduction. It is confirmed in ~\cite{STAR-FC,NDDe} that structural information can improve the precision of pairwise classification. This direct usage of heuristic structural information may not be conducive to generalization to other datasets. So, NASA is designed to extract structural information in a learning way. To balance precision and recall in the NASA learning process, inspired by graph information bottlenecks ~\cite{GIB} and LTR, we optimize the following objective:
\begin{equation}
    max \;  (\sum_{label = 1}^{}P_{ij}^{linkage}  - \sum_{label = 0}^{}P_{ij}^{linkage} )
\end{equation}
The maximization of the former guarantees the recall and the minimization of the latter ensures the precision.

\begin{figure}[tbp]
    \centering
	\includegraphics[width=0.8\linewidth]{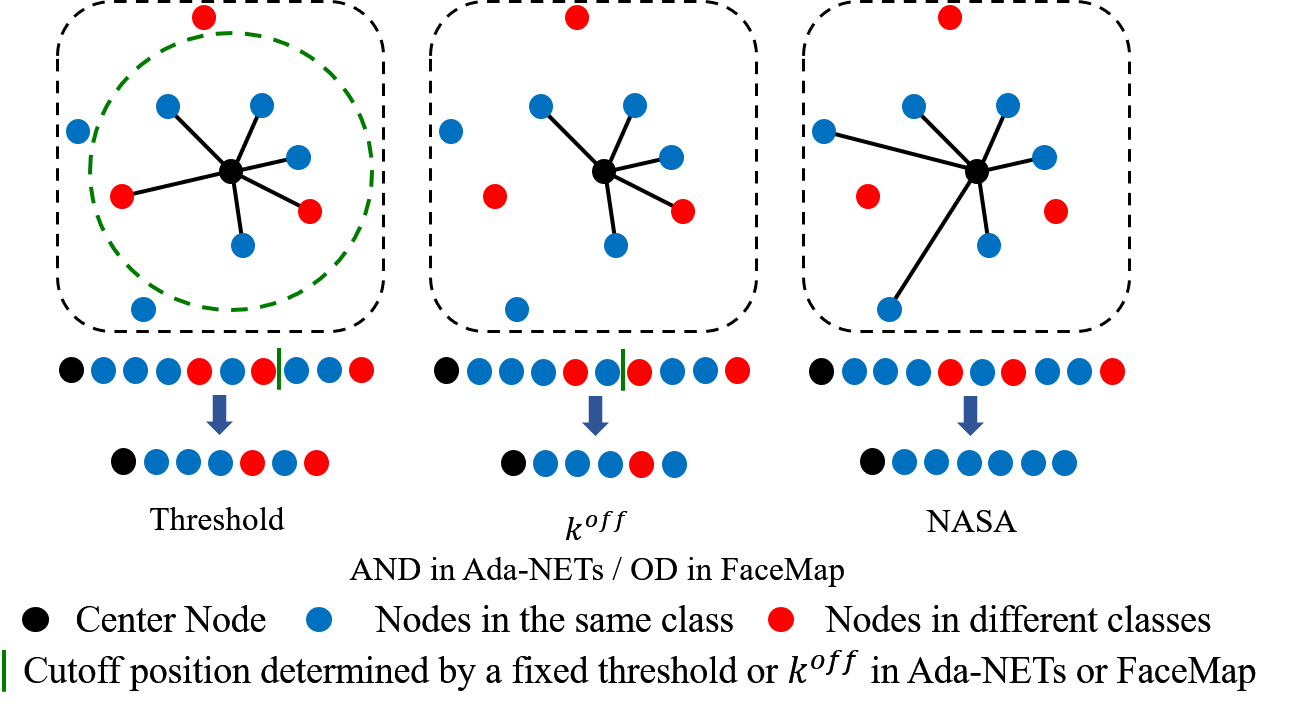}
	\caption{Differences in different methods for constructing subgraphs for face clustering.}
\end {figure}

\section{Methodology}

\subsection{Overview}

Given a node set $V=[v_1,v_2,…,v_N]$ of features $F=[f_1,f_2,…,f_N] \in R^{N \times D}$ extracted from a pre-trained CNN model, where N presents the number of nodes and D is the dimension of each feature. Let $K(v_i)$ be the k-nearest neighbor set for node $v_i$ and $s_{ij}$ be the cosine similarity between node $v_i$ and its neighbor $v_j$. $A_i$ denotes subgraph adjacency matrix of node $v_i$.

A Neighborhood-Aware Subgraph Adjustment method (NASA) is proposed to provide new subgraphs with a higher level of quality for face clustering. Figure 4 displays an overview of the proposed procedure. The entire procedure consists of two steps: (a) train a MLP with enhanced pairwise features and neighborhood embeddings of enclosed subgraphs to adjust subgraphs; (b) perform the clustering via different methods, such as GCN aggregation or random walk.

\begin{figure*}[htbp]
    \centering
	\includegraphics[width=0.75\linewidth]{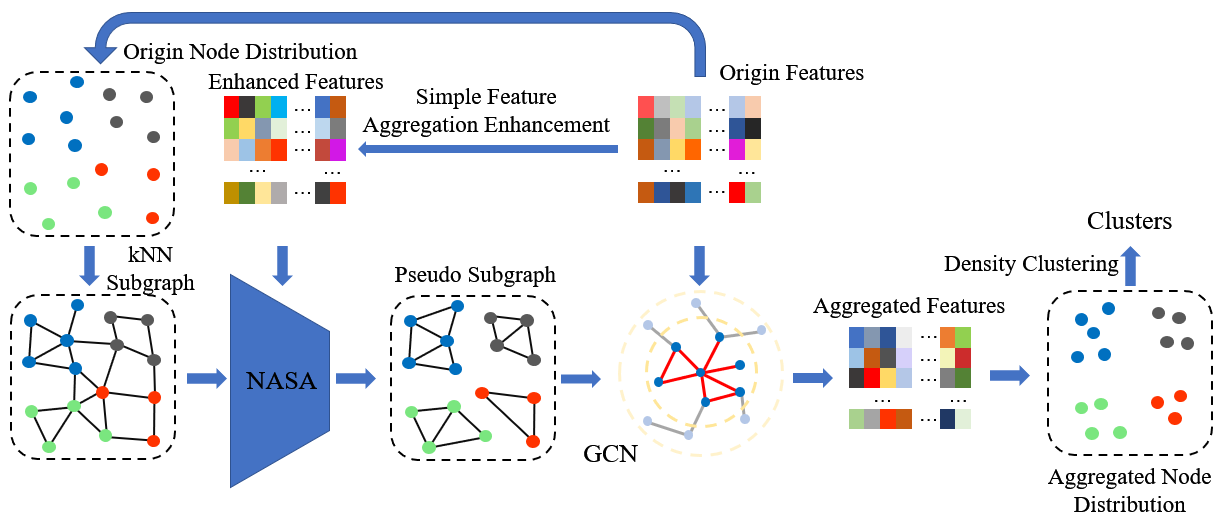}
	\caption{An overview of the whole proposed procedure.}
\end {figure*}

\subsection{Neighborhood-Aware Subgraph Adjustment}

NASA consists of two modules: feature discriminator (FD) and neighborhood discriminator (ND). Numerous earlier studies have shown that local neighborhood structures can improve embedding quality, which is more suitable for pairwise classification or other tasks. Therefore, different structure information is applied in FD and ND respectively. The schematic diagram of NASA is shown in Figure 5.

FD extracts pairwise feature embeddings by concatenating the features of two nodes and feeding them into the MLP. Several researches have demonstrated that the representation can be improved by neighborhood feature propagation. There are various methods for propagation, such as Transformers ~\cite{NDDe}, GCNs ~\cite{DANET} or Mean Aggregation ~\cite{L-GCN}. For efficiency of the proposed FD, a simple mean aggregation approach is introduced as follow:
\begin{equation}
\begin{aligned}
    f{'}_i \; = \; \frac{1}{K{'}} \sum_{j \in K^T(v_i)}^{K{'}}  f_j , \; f \in F^{N \times D} \\
    K^T(v_i) \;= \; \left \{ v_{i_j} \; \mid  \; s_{ij} \ge t_1 , \; v_j \in K(v_i) \right \}
\end{aligned}
\end{equation}
where $f{'}_i$ represent new features and $K^T(v_i)$ is the node set which contains nodes with a higher similarity than threshold $t_1$ in original $K(v_i)$. The original features and new features are concatenated into enhanced features ($f^{enh}$) as follow:
\begin{equation}
    f_{ij}^{enh}\;=\;[\;f_i \; \mid \;  f{'}_i\; \mid \; f_j \; \mid \;  f{'}_j\;]
\end{equation}

Motivated by ~\cite{WLNM,SEAL}, ND incorporates the structural features to represent enclosed subgraphs between two nodes. In order to increase the precision of the enclosed subgraph which represent the connection probability of two nodes, the first-order and second-order neighbors are included. The neighbor selection rule is as follow:
\begin{gather}
    \begin{split}
            V_{ij}^{E1} = \left \{ v_c \mid s_{ic}\ge t_2,c \in K(v_i) \right \} \cup \\
                 \left \{ v_c \mid s_{jc}\ge t_2,c \in K(v_j)\right \} \\
            V_{ij}^{E2} =\left \{ v_d  \;\mid\;  s_{cd}  \;\ge\; t_2, \;d \in K(v_c) \right \}  \\
            V_{ij}^{E} \;=\; \left \{ \; V_{ij}^{E1} \;\mid\; V_{ij}^{E2} \;\right \}
    \end{split}
\end{gather}
where $V_{ij}^{E1}$ and $V_{ij}^{E2}$ represent the first-order and second-order neighbors respectively, and $V_{ij}^{E}$ denotes the node set of the enclosed subgraph between node $i$ and $j$. The selected threshold $t_2$ can be same as $t_1$ or can be 0.

In order to reflect the distinct contribution of different nodes to the linkage probability between two central nodes, it is necessary to label them according to their distances to central nodes. Even though many researches ~\cite{WLNM,SEAL} have demonstrated the excellent performance of the shortest paths between central node and its neighbors, the computation complexity is too high to be extended to large face datasets. The main propose of the shortest paths here are to measure the distances from different nodes to central nodes. Therefore, a simplified way of node labeled approach is adopted, which consists of two parts, and the shortest path distances are replaced by the cosine distances. The first part is related to the order of the neighbors and the central nodes. The order of central nodes is 0. The second part is the similarities between the neighbors and central nodes.  The strategy is as follow:
\begin{equation}
    \begin{aligned}
        dist_{in} \;=\; \left\{\begin{matrix}
        s_{in}, \;\; s_{in} \;\ge\; t_2\\
        dist_{Max}, \;\;    otherwise
        \end{matrix}\right.  \\
        dist_n \;=\; \frac{dist_{in} \;+\; dist_{jn}}{2} \;+\; O_n
    \end{aligned}
\end{equation}
where $O_n$ represents the order of node $n$, and $dist_n$ represents the distance between node $n$ and central nodes ($i$ and $j$). For nodes whose similarity with central nodes is lower than the threshold $t_2$, set its distance to $dist_{Max}$. In practice, we empirically set it as 4. All nodes in enclosed subgraphs will be processed and sorted. The final distance lists can be defined as the structural features $f_{ij}^{str}=[dist_1,dist_2,…,dist_n]$.

For each node in the datasets, let it and its k-nearest neighbors $K(v_i)$ form node pairs one by one. The enhanced features $f^{enh}$ and the structure features $f^{str}$ will be fed to FD and ND respectively. Then, the embeddings of FD and ND are concatenated and the linkage probabilities are predicted as follow:
\begin{equation}
    P_{ij}^{linkage} \;=\; MLP([\;FD(f_{ij}^{enh})  \mid  ND(f_{ij}^{str})])
\end{equation}
The schematic of NASA is depicted in Figure 5. The specific structure of FD and ND will be given in the Appendix. During the test stage, the similarities in the original kNNs will be replaced with probabilities predicted by NASA. 

\begin{figure*}[htbp]
    \centering
	
	\includegraphics[width=0.6\linewidth]{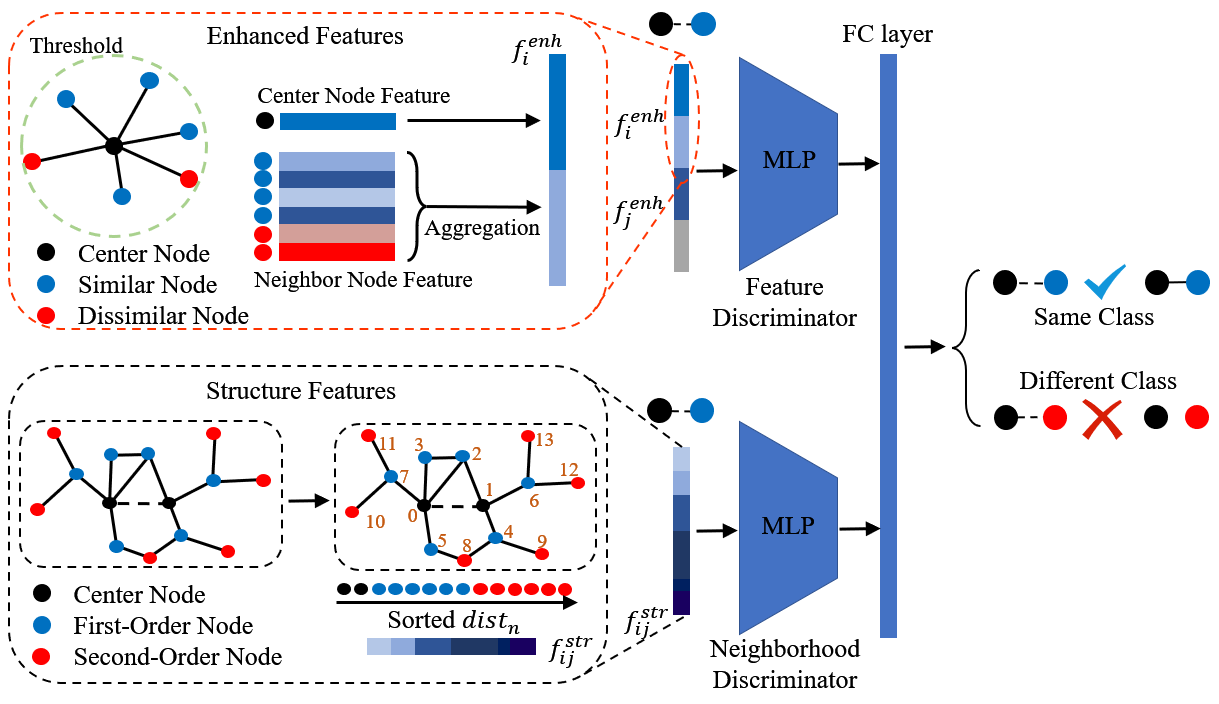}
	\caption{Schematic diagram of the proposed NASA module. Note that colors of the nodes in FD and ND represent different meanings.}
\end {figure*}

\subsection{Face Clustering with Better Subgraphs}

In the previous section, the procedure of NASA is elaborated. Cleaner and richer subgraphs of most nodes can be available for face clustering. 

Previous studies tend to construct subgraphs for all nodes based on cosine kNNs. Here, new subgraphs for clustering are redefined as follow:
\begin{equation}
    \begin{aligned}
        K^P (v_i )\;&=\;  \left \{ v_j  \;\mid\;  P_{ij}^{linkage} \;\ge\; t_3,\;v_j\ \;\in\;K(v_i) \right \} \\
        A_{ij}\;&=\;\left\{\begin{matrix}
                1, \;\; v_j \;\in\; K^P(v_i) \\
                0, \;\;\;\;\;\; otherwise
                \end{matrix}\right.
    \end{aligned}
\end{equation}
where $t_3$ denotes the probability threshold. Since NASA is well trained, the confidences will be polarized. Therefore, the selection of  $t_3$ is more liberal. A variety of clustering methods can be used with the better subgraphs. Here the GCNs are adopted for better generalization.

The GCN layer is defined as follow:
\begin{equation}
    F^{l+1}  \;=\; \sigma  (\widetilde{D}^{-1}\widetilde{A}F^l W^l+F^l W_{skip}^l)
\end{equation}
where $\widetilde{A} =A+I$ and I is an identity matrix. $\widetilde{D}$ denotes the degree matrix whose element is $\widetilde{D}_{ii}= {\textstyle \sum_{j=1}^{n}} \widetilde{A}_{ij}$  .  Each GCN layer is followed by a SELU activation ~\cite{SELU}, which is consistent with ~\cite{GCNFA}. After 2 GCN layers, the features generated are passed through ArcFace Loss proposed in ~\cite{ArcFace}, which can improve the distinguishability of aggregated features. This is not the key here, so there is no need to dwell on other better face recognition loss.

After feature aggregation, the density peak clustering algorithm is applied. Approximate NDDe ~\cite{NDDe} demonstrates a high-quality density measurement approach, and can be adopted here. Each node will connect to nodes with higher density than itself, while a similarity threshold is used to remove the connections that exists between non-identical node pairs. Then, the connected components can be obtained as predicted clusters.

\begin{algorithm}[tb]
    \caption{Train Procedure of the proposed method}
    \label{alg:algorithm}
    \textbf{Input}: Training set $V$; Training features $F \in R^{N \times D}$; Similarity threshold $t_1$; Number of neighbors for constructing subgraphs $k$.\\
    \textbf{Output}: clusters $C$
    \begin{algorithmic}[1] %[1] enables line numbers
        \STATE A = Build Graph($F$, $t_1$, $k$).
        \FOR{node $i$ in $V$}
        \STATE find k-nearest neighbors for node $i$
        \STATE form node pairs for node $i$ and its neighbors
        \STATE construct enhanced features $f_{ij}^{enh}$ for each node pair
        \STATE construct structure features $f_{ij}^{str}$ for each node pair
        \STATE predict $P_{ij}^{linkage}$ for each node pair via (6)
        \STATE replace original cosine similarities with $P_{ij}^{linkage}$
        \ENDFOR
        \FOR{node $i$ in $V$}
        \STATE  construct subgraphs $A_i$ for node $i$
        \STATE aggregate features via GCN($F$, $A_i$) in (8)
        \ENDFOR
        \STATE Obtain the clusters $C$ via approximated NDDe
        \STATE \textbf{return} $C$
    \end{algorithmic}
\end{algorithm}

\section{Experiments}

\subsection{Experimental settings}

\paragraph{Datasets.} Three mainstream clustering benchmark datasets are chosen, MS1M ~\cite{MS1M}, IJB-B ~\cite{IJB-B} and DeepFashion ~\cite{DeepFashion}. MS1M is a widely used face clustering dataset consisting of 5.8M images of 86K individuals. A partial sample was used for comparison in ~\cite{GCN-VE}. The whole dataset is divided into 10 parts. Part0 is used for training and Part1 is used for testing. This work follows this practice. Three different clustering protocols are available for IJB-B, containing 18171, 36575 and 68195 images of 512, 1024 and 1845 individuals, respectively. We follow the settings in ~\cite{L-GCN}, in which the CASIA dataset is used for training and these three different protocols are used for comparisons. DeepFashion is also frequently used for comparisons in some previous studies. 25752 images from 3997 categories are used for training and 26960 images from 3984 categories are used for testing. All features of these datasets are obtained from ~\cite{L-GCN,GCN-VE} for fair comparisons. 

\paragraph{Metrics.} The comparisons are made in different ways on different datasets, following the settings of previous work. Specifically, Pairwise F-score and Bcubed F-score are used to measure model performance on the MS1M and DeepFashion datasets, while on IJB-B, only Bcubed F-score is used. 

\paragraph{Implementation Details.} Our NASA modules consist of 6/8 linear layers for the feature discriminator and 3 linear layers for the neighborhood discriminator. An additional fully-connected layer is used to fuse the features extracted from FD and ND, and output the predictions. The number of neighbors $k$ is 80 for MS1M and CASIA and 10 for DeepFashion (the same as previous work ~\cite{L-GCN,DANET}). The threshold $t_1$ is 0.8 for MS1M, 0.7 for CASIA and 0.925 for DeepFashion. The hop-1 and hop-2 number for constructing the enclosed subgraphs of ND ($k_1$ and $k_2$) is 60 and 10 for MS1M and CASIA, and 5 and 3 for DeepFashion. Other detailed parameters for the model training are given in the Appendix.

\subsection{Comparisons with State-of-the-Art Methods}

Different face clustering baselines are compared with the proposed methods. Traditional methods include K-Means, DBSCAN, HAC and ARO; learning-based methods consist of LGCN, DA-Net, Clusformer, STAR-FC, FaceT, Pair-Cls, Ada-NETs, NDDe \& TPDi and GCN-F\&A. The proposed method can outperform the current SOTA on different datasets, as shown in Table 1, Table 2 and Table 3. Since the training datasets and the test datasets of MS1M and DeepFashion are homogeneous and the feature distribution is consistent, many methods are effective on these two datasets. However, due to the cross-domain distribution of CASIA and IJB-B, some methods with overfitting cases on MS1M and DeepFashion show significant performance degradation, such as Ada-NETs and NDDe \& TPDi. The proposed NASA method, which learns the subgraphs directly from feature and neighborhood perspectives, can reconstruct the subgraphs. The additional structure information alleviates the overfitting that may occur by using the feature only. On IJB-B, the proposed solution can achieve more than 1\% improvement.

\begin{table}
    \centering
    \begin{tabular}{lrr}
        \hline
        Methods  & $F_P$ &  $F_B$ \\
        \hline
        K-Means     &79.21      &81.23  \\
        DBSCAN      &67.93      &67.17  \\
        HAC         &70.63      &70.46  \\
        ARO         &13.60      &17.00  \\
        \hline
        CDP         &75.02      &78.70  \\
        L-GCN       &78.68      &84.37  \\
        GCN-DS      &85.66      &85.52  \\
        GCN-VE      &87.93      &86.09  \\
        DA-NET      &90.60      &-      \\
        Clusformer  &88.20      &87.17  \\
        STAR-FC     &91.97      &90.21  \\
        FaceT       &91.12      &90.50  \\
        Pair-Cls    &90.67      &89.54  \\
        ADA-NETS    &92.79      &91.40  \\
        InfoMap     &93.98      &92.40  \\
        FaceMap     &94.21      &92.55  \\
        NDDe \& TPDi    &93.22  &92.18  \\
        GCN-F\&A    &94.09      &-      \\
        \hline
        GCN         &92.73      &91.17  \\
        NASA-GCN    &\textbf{94.55}      &\textbf{92.99}  \\
        \hline
    \end{tabular}
    \caption{Comparisons on MS1M-Part1 with regard to Pairwise F-Score and Bcubed F-Score.}
    \label{tab:plain}
\end{table}

\begin{table}
    \centering
    \begin{tabular}{lrrr}
        \hline
        Methods  & 512-$F_B$ &  1024-$F_B$ &  1845-$F_B$\\
        \hline
        K-Means     &61.2      &60.3    &60.0\\
        DBSCAN      &75.3      &72.5    &69.5\\
        ARO         &76.3      &75.8    &75.5\\
        \hline
        L-GCN       &83.3      &83.3    &81.4\\
        DA-NET      &83.4      &83.3    &82.8\\
        FaceT       &83.1      &83.3    &82.2\\
        Pair-Cls    &84.4      &83.3    &82.7\\
        GCN-F\&A    &82.6      &83.0    &81.8\\
        \hline
        NASA-GCN    &\textbf{84.5}     &\textbf{84.3}   &\textbf{84.1}\\
        \hline
    \end{tabular}
    \caption{Comparisons on IJB-B with regard to Bcubed F-Score.}
    \label{tab:plain}
\end{table}

\begin{table}
    \centering
    \begin{tabular}{lrr}
        \hline
        Methods  & $F_P$ &  $F_B$ \\
        \hline
        K-Means     &32.86      &53.77\\
        DBSCAN      &25.07      &53.23\\
        HAC         &22.54      &48.70\\
        ARO         &26.03      &53.01\\
        \hline
        CDP         &28.28      &57.83\\
        L-GCN       &28.85      &58.91\\
        GCN-DS      &29.14      &59.11\\
        GCN-VE      &38.47      &60.06\\
        FaceT       &34.82      &61.29\\
        Pair-Cls    &37.67      &62.17\\
        ADA-NETS    &39.30      &61.05\\
        InfoMap     &38.67      &60.48\\
        NDDe \& TPDi    &40.91  &63.61\\
        GCN-F\&A    &46.16      &65.52\\
        \hline
        NASA-GCN    &\textbf{46.80}      &\textbf{65.64}\\
        \hline
    \end{tabular}
    \caption{Comparisons on DeepFashion with regard to Pairwise F-Score and Bcubed F-Score.}
    \label{tab:plain}
\end{table}

To better demonstrate the impact of the proposed approach for GCNs, all nodes with 7 different labels are shown in Figure 6. The t-SNE plots of the original features distribution and different prediction results of L-GCN, GCNs described in Section 4.3, and GCNs with the proposed subgraph adjustment. It can be seen that when training the GCN using only the original subgraphs, a large number of singletons appear. Some nodes may even split out from the main clusters, which leads to the division of the large clusters into different small clusters. Thanks to the addition of more distant nodes, the better subgraphs generated by NASA can pull each other closer and reduce the risk of splitting. Significant reduction in singletons and small clusters can be observed.

\subsection{Comparisons with other Subgraph Adjustment Methods}

The AND module in Ada-NETs and OD module in FaceMap were developed based on a strong assumption on feature distributions, i.e., that nodes with higher cosine similarities are more likely to be the same kind. These methods sort the neighbors by similarities and obtain the cutoff position $k^{off}$ by explicitly fitting the similarity distribution or directly learning from the features. A high risk of overfitting can be observed, which is less effective on hard samples. The same type of nodes with low similarities and non-identical nodes with high similarities cannot be detected correctly. The former ones will be excluded from the subgraphs and the latter will be selected into the subgraphs as noise. Higher precision of the subgraphs can be achieved through these approaches, but they cannot improve recall. The transformer in NDDe \& TPDi, which can also be regarded as a subgraph adjustment scheme, improves the quality of the subgraphs by replacing the original similarities with the confidences of the Transformer’s predictions. However, it only uses features, making it hard to provide high-quality predictions for node pairs in the confusion region (as shown in Figure 1) despite of excellent performance of the Transformer. The results in Table 4 can fully demonstrate the superiority of the proposed NASA approach. (To evaluate the effectiveness of different methods more fairly, the value of $k$ in kNN is set to 80, so that the number of node pairs is the same).

\begin{figure}[htbp]
    \centering
	\begin{minipage}{0.49\linewidth}
	\centering
	\includegraphics[width=3.5cm,height=3.5cm]{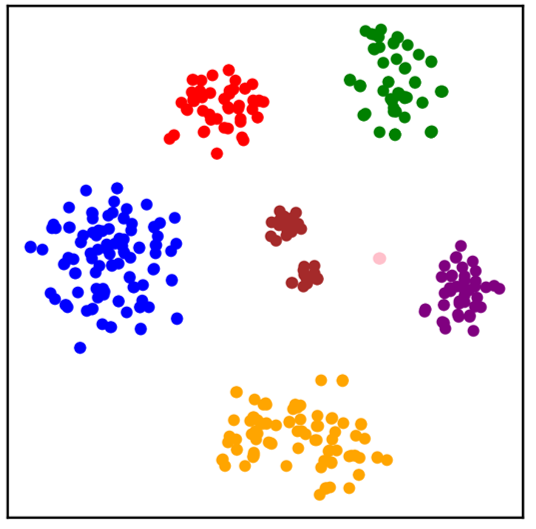}
	\caption*{(a)}
	\label{(a)}
	\end{minipage}
        \begin{minipage}{0.49\linewidth}
	\centering
	\includegraphics[width=3.5cm,height=3.5cm]{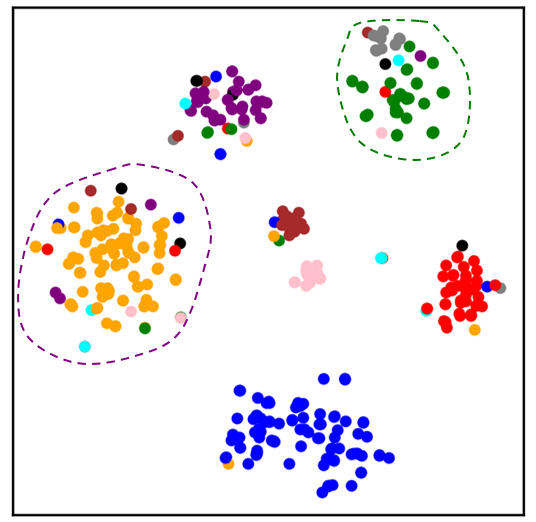}
	\caption*{(b)}
	\label{(b)}
	\end{minipage}

	\begin{minipage}{0.49\linewidth}
	\centering
	\includegraphics[width=3.5cm,height=3.5cm]{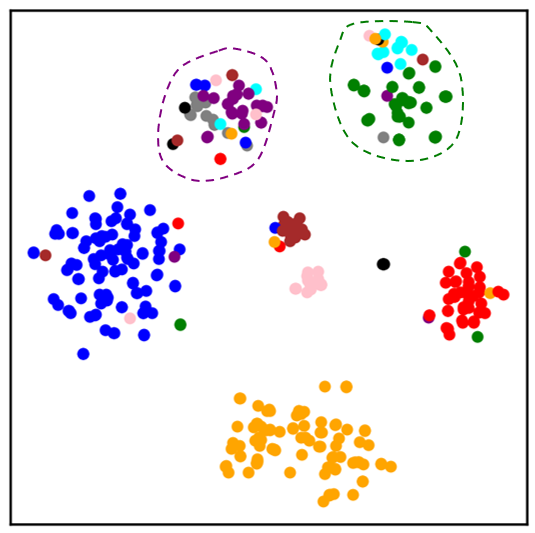}
	\caption*{(c)}
	\label{(c)}
	\end{minipage}
	\begin{minipage}{0.49\linewidth}
	\centering
	\includegraphics[width=3.5cm,height=3.5cm]{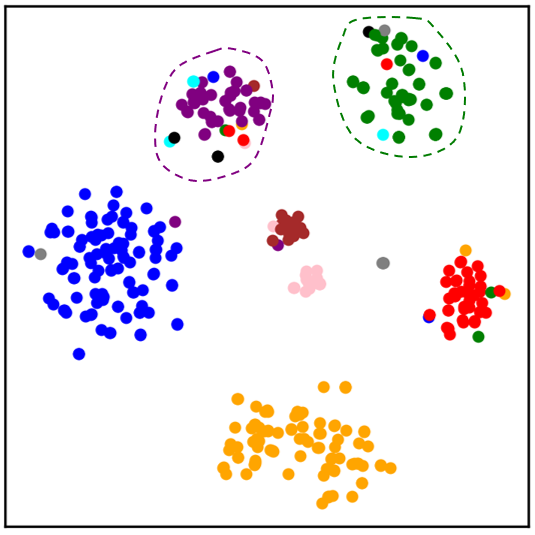}
	\caption*{(d)}
	\label{(d)}
	\end{minipage}
	
	\caption{The feature distribution of 7 labels of samples from MS1M via t-SNE. (a) The ground true. The labels are the same when the colors are the same. (b) Predicted results of L-GCN. (c) Predicted results of the GCNs described in Section 3.3. (d) Predicted results of the proposed NASA-GCN. Clearly, NASA can guide the GCNs to generate better aggregate features and alleviate splitting of singletons and small clusters.}
\end {figure}

\subsection{Extensibility Test}

Different subgraph adjustment approaches are discussed in detail and the effectiveness of our proposed NASA approach has been fully tested with GCNs. In this section, two different clustering schemes, NDDe \& TPDi and InfoMap, will be leveraged to better demonstrate the improvement of the proposed method on subgraph quality. 

\paragraph{NDDe \& TPDi.} The authors of ~\cite{NDDe} propagated distances back to calculate the density of each node which strengthens the fairness of small clusters. Also, they proposed a new distance measure approach based on common neighbors. Here, the probabilities predicted by the feature-based Transformer model are substituted with the probabilities predicted by NASA.

\paragraph{InfoMap.} InfoMap is a widely used and effective non-overlapping community detection algorithm that minimizes the encoding length of different clusters by random walks with restarting. Here, the cosine similarities in the original kNNs are replaced by the probabilities predicted by NASA. 

Table 5 fully demonstrates the significant improvements to both of these clustering methods with the higher-quality subgraphs. The stable improvements over three different datasets also confirm the robustness of the proposed NASA method. Due to the lack of the results of NDDe \& TPDi on IJB-B, the authors retrained it with the parameters used on MS1M. However, its performance is poor with possibly wrong parameters, so we omit it here to avoid misleading readers.

\begin{table}
    \centering
    \begin{tabular}{lrrr}
        \hline
        Setting  & Accuracy & Precision & Recall \\
        \hline
        Threshold (0.8)   &83.41  &95.14  &80.45 \\
        AND (ADA-NETS)   &85.05  &92.73  &85.37  \\
        OD (FaceMap)     &64.22  &92.94  &53.01  \\
        Transformer\\(NDDe \& TPDi)   &85.31  &88.95  &90.39  \\
        \hline
        NASA (\emph{ab1})       &86.06  &87.53  &93.42  \\
        NASA (\emph{ab2})       &96.84  &97.64  &\textbf{97.86}  \\
        NASA (\emph{ab3})       &97.09  &98.79  &97.05  \\
        NASA            &\textbf{97.17}  &\textbf{98.80}  &97.07  \\
        \hline
    \end{tabular}
    \caption{Comparisons of different subgraph adjustment methods on MS1M-Part1.}
    \label{tab:plain}
\end{table}

\begin{table}
    \centering
    \begin{tabular}{lccrr}
        \hline
        Dataset                      & Methods                       & Setting & $F_P$    & $F_B$ \\
        \hline
        \multirow{4}{*}{MS1M}        & \multirow{2}{*}{InfoMap}      & origin  & 93.98      & 92.40    \\
                                     &                               & ours    & \textbf{94.80}      & \textbf{93.18}    \\  \cline{2-5} 
                                     & \multirow{2}{*}{NDDe \& TPDi} & origin  & 93.22      & 92.18   \\
                                     &                               & ours    & \textbf{94.08}      & \textbf{92.51}   \\  \cline{1-5} 
        \multirow{4}{*}{IJB-B-1845}  & \multirow{2}{*}{InfoMap}      & origin  & 51.76      & 77.48   \\
                                     &                               & ours    & \textbf{62.06}      & \textbf{79.35}   \\  \cline{2-5} 
                                     & \multirow{2}{*}{NDDe \& TPDi} & origin  & -      & -   \\
                                     &                               & ours    & 65.16      & 79.56   \\  \cline{1-5} 
        \multirow{4}{*}{DeepFashion} & \multirow{2}{*}{InfoMap}      & origin  & 38.67      & 60.48   \\
                                     &                               & ours    & \textbf{44.33}      & \textbf{64.63}   \\  \cline{2-5} 
                                     & \multirow{2}{*}{NDDe \& TPDi} & origin  & 40.91      & 63.61   \\
                                     &                               & ours    & \textbf{43.05}      & \textbf{64.63}     \\
        \hline
    \end{tabular}
    \caption{Comparisons of the robustness of the proposed NASA via different clustering methods on three mainstream datasets.}
\end{table}

\subsection{Ablation Experiments}

To demonstrate the effect of different modules on pairwise classification and clustering performance, some ablation experiments will be supplemented on MS1M-Part1 datasets. The role of the feature discriminator and neighborhood discriminator with its node label approach is conducted. Different modules are removed gradually in ablation experiments as follows: 
\begin{itemize}
\item[$\bullet$] \emph{ab1}: retain the FD module with its original features.
\item[$\bullet$] \emph{ab2}: add enhanced features to \emph{ab1}.
\item[$\bullet$] \emph{ab3}: add the ND module to \emph{ab2}, but directly replace the proposed node labeling method with node orders in enclosed subgraphs.
\end{itemize}
The numerical results are presented in Table 4. It can be observed that the addition of these modules provides NASA with a performance boost that can generate better subgraphs for subsequent GCNs. 

\section{Conclusions}

This paper proposed an effective face clustering method aiming to relieve the conflict between noise suppression and ground-true preservation in the subgraphs during the aggregation stage of GCNs. A neighborhood-aware subgraph adjustment method is introduced to generate high-quality subgraphs and a new node label approach is proposed to reduce prediction time and improve prediction performance, which effectively improves the representation ability of subsequent aggregated features of GCNs. Experiments demonstrate the effectiveness of the NASA approach and SOTA performance on three mainstream datasets can be successfully achieved. The robustness is also confirmed compared via different clustering methods, which demonstrates excellent scalability.

\appendix
% \setcounter{table}{0}

% \begin{appendices}

\section{The Detailed Results on MS1M}

The detailed results on MS1M are given in Table 6. The kNNs required for the experiments were obtained via faiss-gpu. The kNN for MS1M-Part9 could not be obtained due to the GPU memory limitation.

\begin{table*}[htbp]
    \centering
\begin{tabular}{lcccccccc}
\hline
             & \multicolumn{2}{l}{Part1(584K)} & \multicolumn{2}{l}{Part3(1.74M)} & \multicolumn{2}{l}{Part5(2.89M)} & \multicolumn{2}{l}{Part7(4.05M)} \\ \hline
             & $F_P$          & $F_B$          & $F_P$          & $F_B$           & $F_P$           & $F_B$          & $F_P$           & $F_B$          \\ \hline
K-Means      & 79.21          & 81.23          & 73.04          & 75.20           & 69.83           & 72.34          & 67.90           & 70.57          \\
HAC          & 70.63          & 70.46          & 54.40          & 69.53           & 11.08           & 68.62          & 1.40            & 67.69          \\
DBSCAN       & 67.93          & 67.17          & 63.41          & 66.53           & 52.50           & 66.26          & 45.24           & 44.87          \\
ARO          & 13.60          & 17.00          & 8.78           & 12.42           & 7.30            & 10.96          & 6.86            & 10.50          \\ \hline
CDP          & 75.02          & 78.70          & 70.75          & 75.82           & 69.51           & 74.58          & 68.62           & 73.62          \\
L-GCN        & 78.68          & 84.37          & 75.83          & 81.61           & 74.29           & 80.11          & 73.70           & 79.33          \\
GCN-DS       & 85.66          & 85.52          & 82.41          & 83.01           & 80.32           & 81.10          & 78.98           & 79.84          \\
GCN-VE       & 87.93          & 86.09          & 84.04          & 82.84           & 82.10           & 81.24          & 80.45           & 80.09          \\
Clusformer   & 88.20          & 87.17          & 84.60          & 84.05           & 82.79           & 82.30          & 81.03           & 80.51          \\
Pair-Cls     & 90.67          & 89.54          & 86.91          & 86.26           & 85.06           & 84.55          & 83.51           & 83.49          \\
STAR-FC      & 91.97          & 90.21          & 88.28          & 86.26           & 86.17           & 84.13          & 84.70           & 82.63          \\
Ada-NETs     & 92.79          & 91.40          & 89.33          & 87.98           & 87.50           & 86.03          & 85.40           & 84.48          \\
NDDe \& TPDi & 93.22          & 92.18          & 90.51          & 89.43           & 89.09           & 88.00          & 87.93           & 86.92          \\ \hline
Ours         &\textbf{94.55}  &\textbf{92.99}  &\textbf{91.81}  &\textbf{90.16}   &\textbf{90.34}   &\textbf{88.65}  &\textbf{89.25}   &\textbf{87.46}          \\ \hline
\end{tabular}
\caption{The Detailed results on MS1M.}
\end{table*}

\section{Complexity and Time Consumption}

For each of $N$ nodes, the linkage probabilities with its top-$K$ neighbors need to be calculated, and then aggregated, so the complexity of NASA is $O(NK)$. The entire time on MS1M-Part1 was {\textbf{6.2 minutes}}, with NASA taking 5 minutes and GCN taking 1 minutes, compared with Ada-NETs ~\cite{Ada-NETS} (17.5 minutes) and NDDe-TPDi ~\cite{NDDe} (2 minutes).

The cosine distance is used here for performance and speed trade-off. We have chosen the shortest path distance (a common labeling method, see \cite{SEAL} as a comparison. The current NASA takes {\textbf{5 minutes}} on MS1M-Part1, while the shortest path version of NASA takes 1.28 hours.

\section{The Parameters for Training}

The relevant parameters of the proposed method are given in Table 10. The meanings of the relevant parameters of ArcFace Loss and Approximate NDDe, which are not mentioned in the text, can be found from ~\cite{ArcFace} and ~\cite{NDDe}.

\section{The Structures of NASA}

Let $L$ denote a linear layer and $LBR$ denotes a linear layer followed by a BN layer and a leaky RELU layer. All structures are shown in Table 7, Table 8 and Table 9.

\begin{table}[H]
    \centering
    \begin{tabular}{cc}
    \hline
    \multicolumn{1}{c|}{\begin{tabular}[c]{@{}c@{}}Feature\\ Discriminator\end{tabular}} & \begin{tabular}[c]{@{}c@{}}Neighborhood\\ Discriminator\end{tabular} \\ \hline
    \multicolumn{1}{c|}{$LBR(1024 \rightarrow 512)$}                                     &                                                                      \\
    \multicolumn{1}{c|}{$LBR(512 \rightarrow 512)$}                                      &                                                                      \\
    \multicolumn{1}{c|}{$LBR(512 \rightarrow 512)$}                                      & $LBR(1322 \rightarrow 661)$                                          \\
    \multicolumn{1}{c|}{$LBR(512 \rightarrow 512)$}                                      & $LBR(661 \rightarrow 64)$                                            \\
    \multicolumn{1}{c|}{$LBR(512 \rightarrow 512)$}                                      & $LBR(64 \rightarrow 20)$                                            \\
    \multicolumn{1}{c|}{$LBR(512 \rightarrow 256)$}                                      &                                                                      \\
    \multicolumn{1}{c|}{$LBR(256 \rightarrow 256)$}                                      &                                                                      \\
    \multicolumn{1}{c|}{$LBR(256 \rightarrow 40)$}                                       &                                                                      \\ \hline
    \multicolumn{2}{c}{$L(60 \rightarrow 2)$}                                                                                                                   \\ \hline
    \end{tabular}
    \caption{The structure of NASA applied on MS1M-Part1.}
    \label{tab:my-table}
\end{table}

\begin{table}[H]
    \centering
    \begin{tabular}{cc}
        \hline
        \multicolumn{1}{c|}{\begin{tabular}[c]{@{}c@{}}Feature\\ Discriminator\end{tabular}} & \begin{tabular}[c]{@{}c@{}}Neighborhood\\ Discriminator\end{tabular} \\ \hline
        \multicolumn{1}{c|}{$LBR(2048 \rightarrow 1024)$}                                    &                                                                      \\
        \multicolumn{1}{c|}{$LBR(1024 \rightarrow 1024)$}                                    & $LBR(600 \rightarrow 300)$                                           \\
        \multicolumn{1}{c|}{$LBR(1024 \rightarrow 1024$)}                                    & $LBR(300 \rightarrow 64)$                                            \\
        \multicolumn{1}{c|}{$LBR(1024 \rightarrow 512)$}                                     & $LBR(64 \rightarrow 20)$                                             \\
        \multicolumn{1}{c|}{$LBR(512 \rightarrow 256)$}                                      &                                                                      \\
        \multicolumn{1}{c|}{$LBR(256 \rightarrow 40)$}                                       &                                                                      \\ \hline
        \multicolumn{2}{c}{$L(60 \rightarrow 2)$}                                                                                                                   \\ \hline
    \end{tabular}
    \caption{The structure of NASA applied on IJB-B.}
    \label{tab:my-table}
\end{table}

\begin{table}[H]
    \centering
    \begin{tabular}{cc}
        \hline
        \multicolumn{1}{c|}{\begin{tabular}[c]{@{}c@{}}Feature\\ Discriminator\end{tabular}} & \begin{tabular}[c]{@{}c@{}}Neighborhood\\ Discriminator\end{tabular} \\ \hline
        \multicolumn{1}{c|}{$LBR(1024 \rightarrow 512)$}                                     &                                                                      \\
        \multicolumn{1}{c|}{$LBR(512 \rightarrow 512)$}                                      & $LBR(42 \rightarrow 21)$                                          \\
        \multicolumn{1}{c|}{$LBR(512 \rightarrow 512)$}                                      & $LBR(21 \rightarrow 64)$                                            \\
        \multicolumn{1}{c|}{$LBR(512 \rightarrow 256)$}                                      & $LBR(64 \rightarrow 20)$                                             \\
        \multicolumn{1}{c|}{$LBR(256 \rightarrow 256)$}                                      &                                                                      \\
        \multicolumn{1}{c|}{$LBR(256 \rightarrow 40)$}                                       &                                                                      \\ \hline
        \multicolumn{2}{c}{$L(60 \rightarrow 2)$}                                                                                                                   \\ \hline
    \end{tabular}
    \caption{The structure of NASA applied on DeepFashion.}
    \label{tab:my-table}
\end{table}

\begin{table*}[htbp]
\centering
\begin{tabular}{ccccccccc}
\hline
                       &                                                                                &                           & \multicolumn{2}{c}{MS1M}                                    & \multicolumn{2}{c}{IJB-B}                                   & \multicolumn{2}{c}{DeepFashion}                                  \\ \hline
                       &                                                                                &                           & Train                         & Test                        & Train                         & Test                        & Train                           & Test                           \\ \hline
\multirow{15}{*}{NASA} & FD                                                                             & $t_1$                     & 0.8                           & 0.8                         & 0.7                           & 0.7                         & 0.925                           & 0.925                          \\ \cline{2-9} 
                       & \multirow{4}{*}{ND}                                                            & number of hop-1 neighbors $k_1$ & 60                            & 60                          & 60                            & 60                          & 5                               & 5                              \\
                       &                                                                                & number of hop-2 neighbors $k_2$ & 10                            & 10                          & 10                            & 10                          & 3                               & 3                              \\
                       &                                                                                & $t_2$                     & 0.0                           & 0.0                         & 0.7                           & 0.7                         & 0.925                           & 0.925                          \\
                       &                                                                                & $dist_{Max}$             & 0                             & 0                           & 4                             & 4                           & 4                               & 4                              \\ \cline{2-9} 
                       & \multirow{10}{*}{\begin{tabular}[c]{@{}c@{}}general\\ parameters\end{tabular}} & number of neighbors $k$                        & 80                            & 80                          & 80                            & 80                          & 10                              & 10                             \\
                       &                                                                                & seed                      & \multicolumn{6}{c}{42}                                                                                                                                                                       \\
                       &                                                                                & optimizer                 & \multicolumn{6}{c}{SGD}                                                                                                                                                                      \\
                       &                                                                                & learning rate             & \multicolumn{6}{c}{0.1}                                                                                                                                                                      \\
                       &                                                                                & momentum                  & \multicolumn{6}{c}{0.9}                                                                                                                                                                      \\
                       &                                                                                & weight decay             & \multicolumn{6}{c}{0.0001}                                                                                                                                                                   \\
                       &                                                                                & batchsize               & 128                           & 64                          & 128                           & 64                          & 256                             & 128                            \\
                       &                                                                                & epoch                     & 10                            & -                           & 10                            & -                           & 200                             & -                              \\
                       &                                                                                & learning rate strategy    & \multicolumn{6}{c}{\begin{tabular}[c]{@{}c@{}}Every 5 epochs is a cycle.\\ The learning rate is multiplied by 0.1 \\ after first three epochs.\\ It is reset after the cycle.\end{tabular}}  \\
                       &                                                                                & dropout                   & 0.1                           & -                           & 0.2                           & -                           & 0.1                             & -                              \\ \hline
\multirow{16}{*}{GCN}  & \multirow{11}{*}{\begin{tabular}[c]{@{}c@{}}general\\ parameters\end{tabular}} & number of neighbors $k$                        & 80                            & 40                          & 80                            & 80                          & 10                              & 10                             \\
                       &                                                                                & $t_3$                        & 0.8                           & 0.8                         & 0.9                           & 0.8                         & 0.4                             & 0.4                            \\
                       &                                                                                & seed                      & \multicolumn{6}{c}{42}                                                                                                                                                                       \\
                       &                                                                                & optimizer                 & \multicolumn{6}{c}{SGD}                                                                                                                                                                      \\
                       &                                                                                & learning rate             & \multicolumn{6}{c}{0.01}                                                                                                                                                                     \\
                       &                                                                                & momentum                  & \multicolumn{6}{c}{0.9}                                                                                                                                                                      \\
                       &                                                                                & weight decay             & \multicolumn{6}{c}{0.0001}                                                                                                                                                                   \\
                       &                                                                                & batchsize               & 512                           & 256                         & 512                           & 256                         & 256                             & 128                            \\
                       &                                                                                & epoch                     & 20                            & -                           & 10                            & -                           & 120                             & -                              \\
                       &                                                                                & learning rate strategy    & \multicolumn{6}{c}{\begin{tabular}[c]{@{}c@{}}Every 10 epochs is a cycle.\\ The learning rate is multiplied by 0.1 \\ after 2nd, 5th, 8th epoch.\\ It is reset after the cycle.\end{tabular}} \\
                       &                                                                                & dropout                   & 0.1                           & -                           & 0.1                           & -                           & 0.2                             & -                              \\ \cline{2-9} 
                       & \multirow{2}{*}{\begin{tabular}[c]{@{}c@{}}ArcFace\\ Loss\end{tabular}}        & s                         & \multicolumn{6}{c}{40}                                                                                                                                                                       \\
                       &                                                                                & m                         & 0.25                          & -                           & 0.0                           & -                           & 0.0                             & -                              \\ \cline{2-9} 
                       & \multirow{3}{*}{\begin{tabular}[c]{@{}c@{}}Approximate\\ NDDe\end{tabular}}    & k                         & -                             & 90                          & -                             & 80                          & -                               & 11                             \\
                       &                                                                                & sigma                     & -                             & 0.1                         & -                             & 0.1                         & -                               & 0.02                           \\
                       &                                                                                & Max Connection            & -                             & 1                           & -                             & 80                          & -                               & 1                              \\ \hline
\end{tabular}
\caption{The parameters used in the proposed methods.}
\label{tab:my-table}
\end{table*}

% \end{appendices}

%% The file named.bst is a bibliography style file for BibTeX 0.99c
\bibliographystyle{named}
\bibliography{ijcai22}

\end{document}